\documentclass{article}

\usepackage{microtype}
\usepackage{graphicx}
\usepackage{subfigure}
\usepackage{booktabs} 

\usepackage{amsmath}
\usepackage{amssymb}

\usepackage{pifont}
\usepackage{natbib}
\usepackage{hyperref}

\global\long\def\th{\boldsymbol{\theta}}%
\global\long\def\x{\boldsymbol{x}}%
\global\long\def\a{\boldsymbol{a}}%
\global\long\def\L{\mathcal{L}}%
\global\long\def\D{\mathcal{D}}%
\global\long\def\E{\mathbb{E}}%

\usepackage[accepted]{mlsys2020}

\mlsystitlerunning{Adaptive Dense-to-Sparse Paradigm for Pruning Online Recommendation System with Non-Stationary Data}

\begin{document}

\twocolumn[
\mlsystitle{Adaptive Dense-to-Sparse Paradigm for Pruning Online Recommendation System with Non-Stationary Data}



\mlsyssetsymbol{equal}{*}

\begin{mlsysauthorlist}
\mlsysauthor{Mao Ye}{to,fb}
\mlsysauthor{Dhruv Choudhary}{fb}
\mlsysauthor{Jiecao Yu}{fb}
\mlsysauthor{Ellie Wen}{fb}
\mlsysauthor{Zeliang Chen}{fb}
\mlsysauthor{Jiyan Yang}{fb}
\mlsysauthor{Jongsoo Park}{fb}
\mlsysauthor{Qiang Liu}{to}
\mlsysauthor{Arun Kejariwal}{fb}
\end{mlsysauthorlist}
\graphicspath{ {./figure/} } 

\mlsysaffiliation{to}{University of Texas, Austin}
\mlsysaffiliation{fb}{Facebook Inc}
\mlsyscorrespondingauthor{Mao Ye}{my21@cs.utexas.edu}

\mlsyskeywords{Machine Learning, MLSys}

\vskip 0.3in

\begin{abstract}
Large scale deep learning provides a tremendous opportunity to improve the quality of content recommendation systems by employing both wider and deeper models, but this comes at great infrastructural cost and carbon footprint in modern data centers. Pruning is an effective technique that reduces both memory and compute demand for model inference. However, pruning for online recommendation systems is challenging due to the continuous data distribution shift (a.k.a non-stationary data). Although incremental training on the full model is able to adapt to the non-stationary data, directly applying it on the pruned model leads to accuracy loss. This is because the sparsity pattern after pruning requires adjustment to learn new patterns.
To the best of our knowledge, this is the first work to provide in-depth analysis and discussion of applying pruning to online recommendation systems with non-stationary data distribution. Overall, this work makes the following contributions: 1) We present an adaptive dense to sparse paradigm equipped with a novel pruning algorithm for pruning a large scale recommendation system with non-stationary data distribution; 2) We design the pruning algorithm to automatically learn the sparsity across layers to avoid repeating hand-tuning, which is critical for pruning the heterogeneous architectures of recommendation systems trained with non-stationary data.


\end{abstract}
]

\printAffiliationsAndNotice{}


\section{Introduction}
Large scale deep neural networks have achieved groundbreaking success in various cognitive and recommendation tasks including, but not limited to, image classification~\citep{he2016deep}, speech recognition~\citep{amodei2016deep}, and machine translation~\citep{wu2016google}. However, modern deep neural networks have blown up to hundreds of billions or even trillions of parameters, e.g., GPT-3~\citep{brown2020language}, Turing-NLG~\citep{deepspeed}, Megatraon-LM~\citep{shoeybi2020megatronlm}, and thus training and serving these models requires massive energy consumption that ultimately translates to carbon emissions~\citep{carbon-footprint}. It is estimated that modern data centers consume 200 TW~\citep{energy-demand-1, energy-demand-2} of power every year, which accounts for 1\% of global electricity usage and is expected to increase to 20\% by 2030. A large part of this demand is fuelled by modern deep learning systems, with inference cost dominating (80-90\%) the total cost of successful deployment~\citep{algorithmic-efficiency}. Hence, a number of model compression techniques~\citep{gupta2020compression} such as pruning, quantization~\citep{quantization-1}, and distillation~\citep{distillation-1} have been developed and deployed.

\vspace{-0.1cm}
In recent years, there has been a similar surge~\citep{naumov2019deep, deep-wide, din, ncf} in the development of large scale deep recommendation networks as well. Personalization and content recommendation systems have become ubiquitous on both edge and data center scale systems, and thus, scaling these models poses an even steeper demand on modern infrastructure. Unlike other domains, recommendation models present a new set of challenges due to data non-stationarity, and hence model compression techniques such as network pruning need to be re-evaluated in this context.

\vspace{-0.1cm}
Recommendation models usually employ wide-and-deep~\citep{deep-wide} fully-connected (FC) layers. Recent work~\citep{park2018deep} has shown the modern data centers can spend as much as 42\% of total computation in fully-connected (FC) layers during the serving of recommendation systems. Unstructured network pruning is an effective technique that has been shown to reduce the computation cost of FC layers. For example, with 90\% of parameters in FC layers zeroed out, the computation cost can be reduced by 2-3x~\citep{wang2020sparsert}. Meanwhile, the memory footprint needed to hold the models in memory~\citep{gupta2020architectural} and the required communication memory bandwidth can also be reduced. This is particularly important in production systems, which typically load a large number of models simultaneously in memory~\citep{park2018deep}. 

Pruning the recommendation system is advantageous but has a number of challenges:
\vspace{-0.3cm}
\begin{itemize}
  \item Online recommendation systems are lifelong learners \citep{lifelong-learning} and they need to keep adapting to a non-stationary data distribution using an incremental training paradigm. The non-stationarity is rooted in the flux in content (videos, movies, text, etc.) that gets continuously added/removed from the system. Such data distribution shift results in continuous change in feature distribution and, therefore, the importance of the model parameters. However, existing techniques apply network pruning in a stationary domain and hence these techniques cannot be applied out-of-the-box.
\vspace{-0.1cm}
  \item The architecture of the online recommendataion models is highly heterogeneous. It contains very large embedding tables~\citep{naumov2019deep} to represent categorical features. Besides, various MLPs are deployed to process the dense features and learn the interaction between input features. The FC layers in those MLPs have different propensities to pruning and usually a lot of hand-tuning is required to achieve optimal pruning sparsity to avoid accuracy degradation.
\end{itemize}

\begin{figure} \label{fig: data shift hurt}
\center
\includegraphics[width=0.45\textwidth]{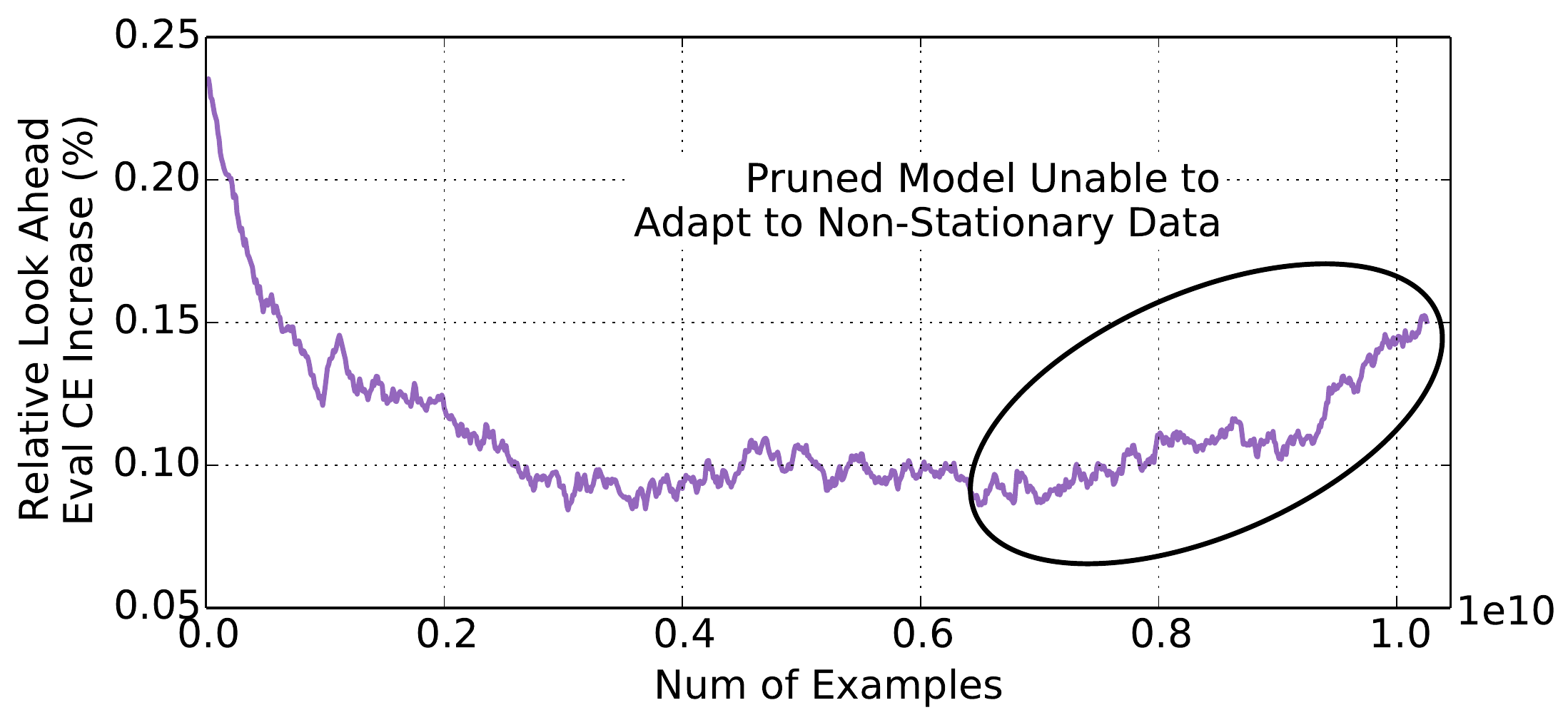}
\vspace{-5mm}
\caption{Relative window loss increase for an 80\% pruned model compared to the full model. The pruned model is not able to adapt to the non-stationary data by directly applying the incremental training as its relative window loss increase significantly rises for later examples.}
\vspace{-4mm}
\label{fig:distribution-shift}
\end{figure}

Incremental training periodically produces a new model by training the previous snapshots with the latest data, and in this way the new model captures the latest data distribution. However, the efficacy of the incremental training requires that the network is sufficiently over-parameterized so that it quickly captures the new data patterns. It has been shown \citep{gordon_2019}, compared with the full dense network, the pruned network is no longer sufficiently over-parameterized and thus is not able to learn as well. Figure \ref{fig:distribution-shift} provides an example showing that a pruned model is no longer be able to adapt to the data distribution shift with incremental training.


We overcome the data distribution shift issue in pruning with a two-fold strategy: a new incremental training paradigm called Dense-to-Sparse (D2S) and a novel pruning algorithm that is specifically designed for it:
\begin{itemize}
  \vspace{-0.3cm}
  \item We propose the D2S paradigm that maintains a full dense model, which is used to adapt to the data non-stationarity by applying incremental training, and periodically produces a pruned model from the latest dense model. It requires a pruning algorithm that is able to produce an accurate pruned model from a well-trained full model with only a limited amount of data for fine-tuning.
  \vspace{-0.2cm}
  \item To satisfy the requirements from the D2S paradigm, we propose a binary auxiliary mask based pruning algorithm. It draws a connection between the heuristic Taylor approximation based pruning algorithm \citep{molchanov2016pruning, molchanov2019importance} and the sparse penalty based methods \citep{he2017channel, liu2017learning}. Based on this connection, the proposed algorithm provides a unified framework that inherits the advantages of both methods. Because of this, the algorithm also automatically learns the sparsity for each model layer and does not depend on heuristics to tune sparsity.
  \vspace{-0.2cm}
  \item We also discuss system design considerations for applying pruning to production scale recommendation systems.
\end{itemize}

\section{Background}

\subsection{Recommendation Models}

\begin{figure}
\center
\includegraphics[width=0.38\textwidth]{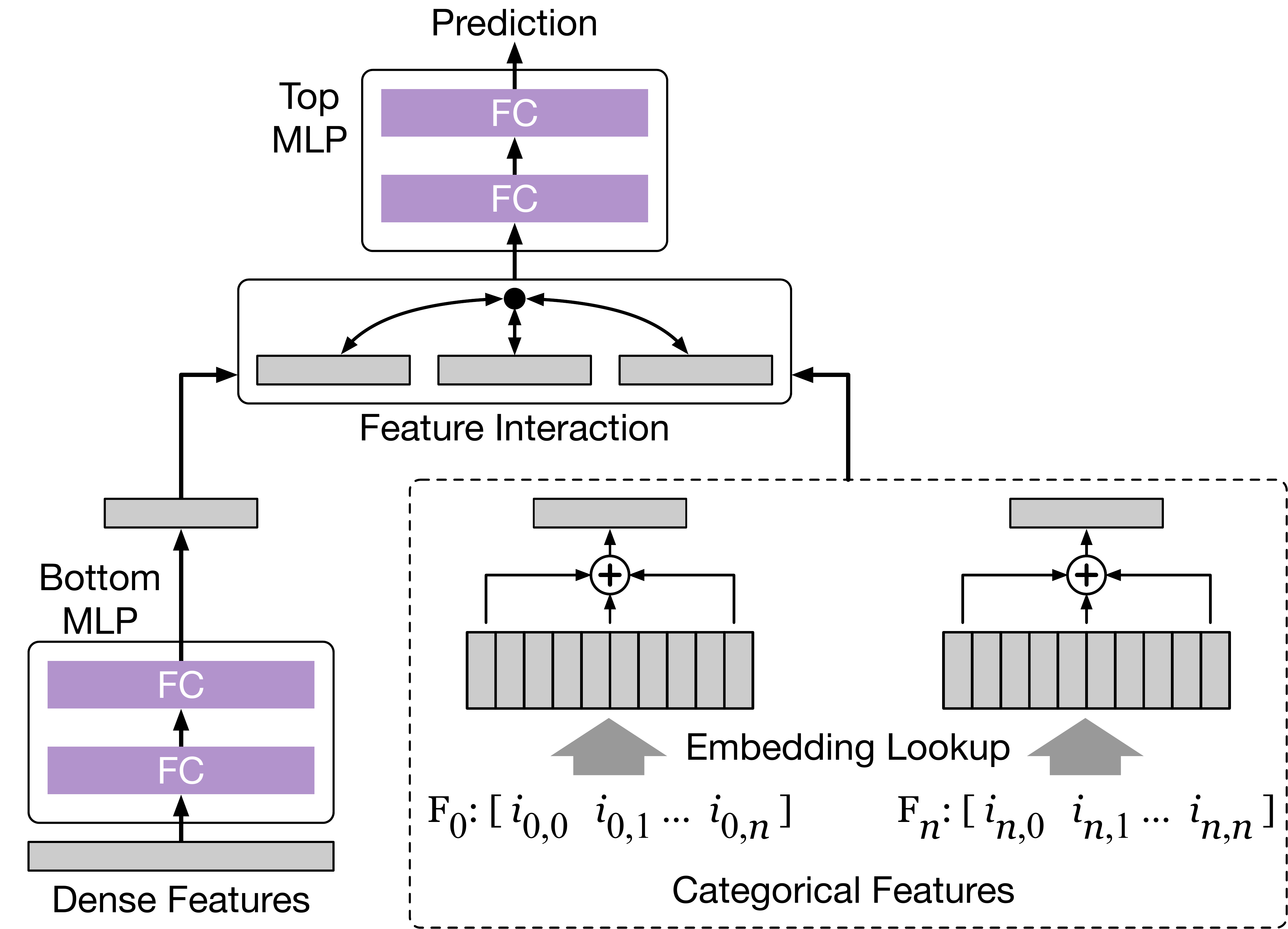}
\vspace{-3mm}
\caption{Recommendation model architecture with dense and categorical features.}
\label{fig:dlrm_diagram}
\vspace{-5mm}
\end{figure}

Figure~\ref{fig:dlrm_diagram} depicts a generalized architecture used in many of the modern recommendation systems~\citep{naumov2019deep, deep-wide, din, ncf}. Personalization models aim to find the most relevant content for a user and employs a combination of real-valued and categorical features. Categorical features are represented using embedding lookup tables where each unique entity (e.g., videos/movies/text) is allocated a row in the table. For multiple entities, the embedding vectors are looked up and pooled together using some aggregate statistic to produce a single embedding for each categorical feature. Dense features are fed directly into Multi-Layer Perceptron (MLP) layers. All the embedding feature vectors coupled with dense features undergo a feature interaction step where combined factors are learned between each pair of embeddings such as Factorization Machines~\citep{fm}, xDeepFM\citep{Lian_2018}. The most common type of interaction is the dot product, as described in the Deep Learning Recommendation Model~\citep{naumov2019deep}(DLRM). In this paper, we consider industrial scale production models similar to the aforementioned architectures.

\vspace{-0.3cm}
\subsection{Data Non-Stationarity}
In contrast to the classical classification/regression problem in which we assume a fixed data population, online recommendation models observe gradually shift data distribution. Suppose at time $t$, the distribution of features and labels $(\boldsymbol{x}, y)$ follows the data distribution $\D_{t}$, and $\D_{t}$ changes gradually with $t$. Given the neural network $f_{\th}$ parameterized by ${\th}$, we define the loss over a data distribution $\D$ as $\L_{\D} \left[f_{\th}\right]=\E_{(\x,y)\sim\D}\ell\left[f_{\th}(\x),y\right]$. As the data is non-stationary, the parameter of the model also needs adjustment in order to capture the data distribution shift. Suppose we set the parameter $\th_t$ at time $t$, our goal is to find $\th_t$ (i.e. the parameters of network at every time $t$) such that the averaged loss across time period $\left[0, T\right]$
\begin{eqnarray} \label{eqn: loss1}
\int_{t\in[0,T]}\L_{\D_{t}}[f_{\th_{t}}]dt
\end{eqnarray}
is minimized. Here $[0, T]$ can be a very long period, e.g., months for a recommendation system in practice. In practice, these systems are optimized for a moving window of average loss because prediction accuracy on new data~\citep{incremental-1} is much more valuable than lifetime accuracy of the model.

\subsection{Incremental Training} \label{fig: increment}
$\th_t$ can be learned with gradient descent with training data. However, in practice, due to the need of deploying the recommendation model for serving, it is hard to update the parameters in real time~\citep{incremental-2, incremental-1}. Instead, the training process is discretized, where parameters are updated at certain fixed time points using the accumulated batch of data. Without loss of generality, suppose that the parameters are only updated at $\mathcal{T}\in[0,T]$ and we assume the time points for updates are equally distributed, i.e., 
\[
\mathcal{T}=\{0,\Delta,2\Delta,....,T-\Delta\}
\]
and thus $\cup_{t\in\mathcal{T}}[t,t+\Delta]=[0,T]$ where $\Delta$ denotes the length between two consecutive time points for model update.

Instead of using the loss in Equation~\ref{eqn: loss1}, $\th_t$ is learned via minimizing the following loss
\begin{eqnarray} \label{eqn: loss2}
\sum_{t\in\mathcal{T}}\int_{s\in[t,t+\Delta]}\L_{\D_{s}}[f_{\th_{t}}]ds.
\end{eqnarray}
The incremental training is conducted in the following way. We start with a pre-trained model $f_{\th_0}$, which is trained with offline data, at time $t=0$. To obtain a new model at $t=k\Delta$ ($k$ is a positive integer), we start with the previous snapshot $f_{\th_{(k-1)\Delta}}$ and apply gradient descent on this previous snapshot with the data collected between time $[(k-1)\Delta, k\Delta)$. This latest model $f_{\th_{k\Delta}}$ will be fixed and deployed for serving between time $k\Delta$ and $(k+1)\Delta$.

\begin{figure}
\center
\includegraphics[width=0.45\textwidth]{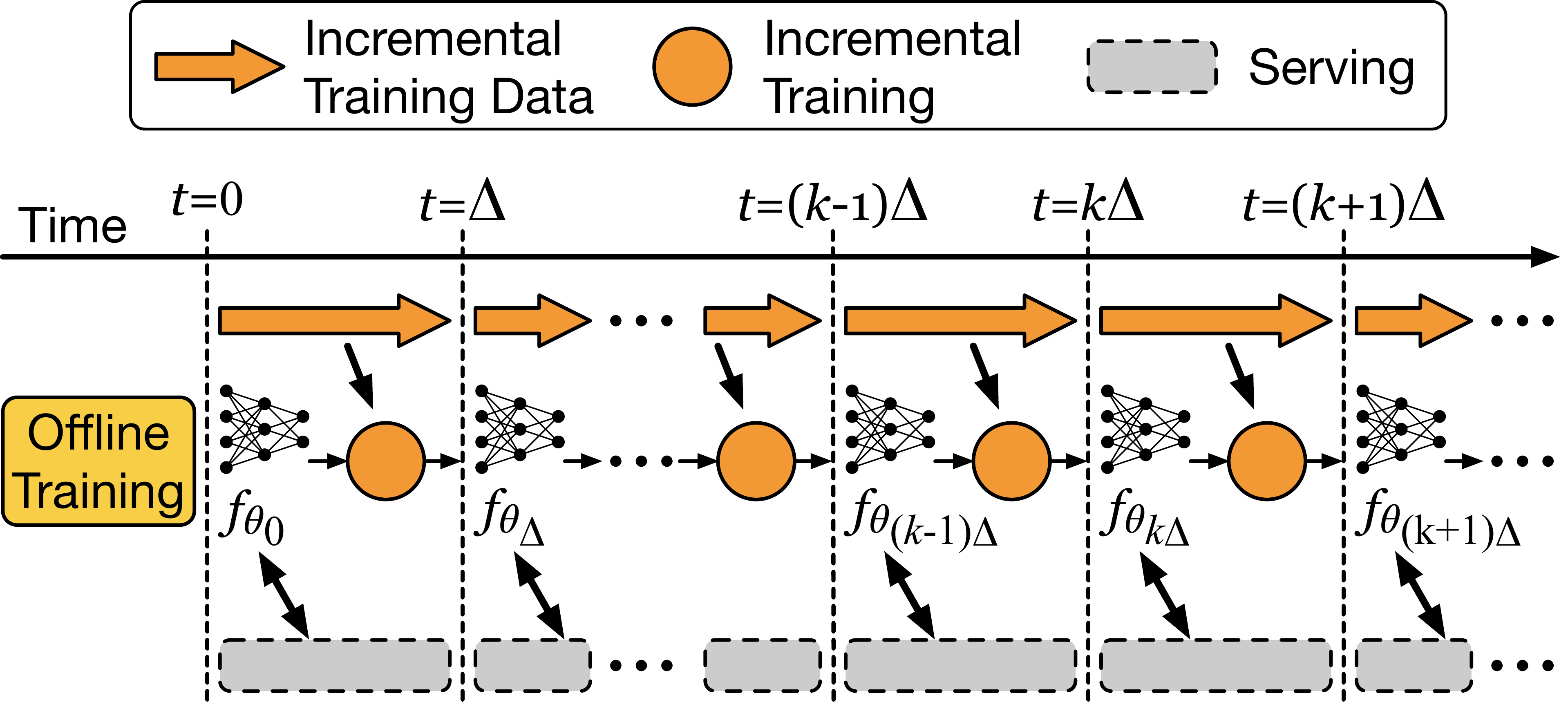}
\vspace{-3mm}
\caption{Incremental training paradigm.}
\label{fig:incremental_training}
\vspace{-5mm}
\end{figure}

Notice that the data distribution shift is a slow and gradual process and thus once $\Delta$ is not too large, the data distribution between $t$ and $t+\Delta$ is similar and thus the model $f_{\th_{k\Delta}}$ is able to give accuracy prediction in the coming time window $[k\Delta, (k+1)\Delta]$.

\paragraph{Pass the training data only for a few epochs}
Different from traditional training of neural networks, in which the optimizer passes the training set for a large number of epochs, in each iteration of the online incremental training (i.e., from $t=k\Delta$ to $t=(k+1)\Delta$), the model will be severely over-fitted if the optimizer passes the training data generated between $[k\Delta,(k+1)\Delta]$ for too many epochs~\citep{zheng2020shadowsync}. This is similar to the catastrophic forgetting~\citep{kirkpatrick2017overcoming} in lifelong learning. Besides, the system requires the trainers to finish training in $\Delta$ times in order to deploy on time, which also puts constraints on the number of epochs for training.

\subsection{Network Pruning}

Pruning is one of the most important techniques for deep neural network (DNN) model compression. By removing redundant weights and their related computation, pruning can dramatically reduce the computation and storage requirements of large-size DNN models~\cite{zhang2018systematic}. Previous works~\cite{park2016faster,wang2020sparsert} demonstrate that, on general-purpose processors (e.g., CPUs and GPUs), pruning offers significant speedup during DNN model inference. As an example, \citet{wang2020sparsert} shows that a speedup of up to 2.5x can be achieved on Nvidia V100 GPU with an unstructured sparsity of 90\%. Considering CPUs, which have much lower hardware parallelism than GPUs, are widely deployed for running deep learning workloads in data centers~\cite{park2018deep}, an even higher performance benefit is expected~\cite{yu2017scalpel} for pruning on recommendation models.




Learning a sparse network can be formulated by the following constrained optimization problem
\vspace{-2mm}
\begin{eqnarray} \label{opt: constraint}
\min_{\th_{t},t\in\mathcal{T}}\sum_{t\in\mathcal{T}}\int_{s\in[t,t+\Delta]}\L_{\D_{s}}[f_{\th_{t}}]ds\ \ \text{s.t.}\ \left\Vert \th_{t}\right\Vert _{0}\le c\ \forall t\in\mathcal{T}.
\end{eqnarray}
\vspace{-3mm}

where $c$ is the number of non-zero parameters we can have. Notice that the above problem is highly non-convex and empirically directly training a sparse network is not able to achieve high accuracy. Instead, starting with a large network and learning a sparse network via pruning is much better. Also see \citet{liu2018rethinking,ye2020good} for more empirical and theoretical results.

Starting with a large dense neural network, network pruning algorithms remove the redundant parameters before, during, or after the training of the full network. For example, pruning during training means the pruning happens during the training of the large network. The learned sparse network requires fine-tuning or retraining with a certain amount of data in order to obtain good accuracy. We now give a detailed review of different types of pruning algorithms.

\section{Related Work} \label{sec: related}
Model compression has received a lot of attention in the last few years with techniques like Quantization \citep{quantization-1, quantization-2}, Distillation \citep{distillation-1, distillation-2, distillation-3}, and Factorization \citep{factorization-1} making their way to most state-of-the-art models. Network pruning has also been proposed in a number of different settings, and can generally be classified into three classes based on when pruning is applied:

\textbf{Pruning Before Training} Pruning deep learning models before the training process is a recent trend of pruning algorithms. The lottery ticket hypothesis~\cite{frankle2018lottery} demonstrates that particular sub-networks (winning tickets) can be trained in isolation and reach the test accuracy comparable to the original network in a similar number of iteration. \citet{dettmers2019sparse, van2020single,wang2020picking} propose various criteria and techniques to find the winning tickets using only a few batches of training data. However, pruning before training algorithms are not applicable for the online recommendation system. The optimal sparsity structure keeps changing due to the non-stationary data and thus the winning tickets selected using the earlier data might not be a winning ticket for later data.


\textbf{Pruning During Training}
Typical pruning during training algorithms are sparse penalty based, which introduce sparse penalty into the training objective to enforce the sparsity \citep{he2017channel, liu2017learning, ye2018rethinking, wen2016learning, zhou2016less, lebedev2016fast, alvarez2016learning, gordon2018morphnet,  yoon2017combined}. Besides, various methods~\cite{dettmers2019sparse, ding2019global, evci2020rigging} are proposed to keep a sparse model through the training process. Recent works \citep{kusupati2020soft, ltp} learn the threshold for pruning the weight and thus automatically also learn the sparsity per layer. Despite the achieved empirical success and the ability to learn the sparsity, they cannot be applied to our online recommendation models. This is because, as we discuss in Section~\ref{sec:difficulties}, for online recommendation systems, sparse models learn slower than full dense models. Therefore, using these pruning during training methods will prevent the model from capturing the data distribution shift in the long term. 


\textbf{Pruning After Training}
Most pruning algorithm that prune after training use heuristic criteria to measure the weight/neuron importance~\citep{liu2018rethinking, blalock2020state,molchanov2016pruning,li2016pruning,molchanov2019importance}. However, previous heuristic-based methods usually require strong prior-knowledge on the desired sparsity structure, i.e., the desired sparsity for different layers. However, hand-crafting those hyper-parameters can hardly gives an optimal solution, and the solution might also change due to the data non-stationarity. Besides, previous works only consider using a single criterion to measure weight importance. Our algorithm design improves over those methods by automatically learning the sparsity structure and combining multiple criteria to measure weight importance. 

Reconstruction error based methods~\citep{he2017channel,luo2017thinet,zhuang2018discrimination} are also popular pruning after training algorithms. However, those methods are typically designed for structured pruning, and it is unknown how it generalizes to unstructured pruning.

\section{Dense-to-Sparse Paradigm}

\subsection{Difficulties in Incremental Training of Sparse Network}
\label{sec:difficulties}
The success of the incremental training procedure relies on the fact that the full model is able to converge to a good local optimal in every incremental training step and thus captures the distribution shift \emph{in long term}. However, this is not the case for the pruned model. The reasons are two-fold:
\begin{itemize}
    \item First, as the sparse network has much fewer parameters than the full network, in each incremental training step, the sparse network learns slower than the full network and converges to a worse local optimal. This is because the sparse network cannot tap into the benefit from the over-parameterization for optimization~\citep{jacot2018neural,du2018gradient,mei2018mean}. 
    \item Second, the different data distribution at different times might require sparse networks with different sparsity structures, i.e., different layers might need to have different sparsity, or the pruned parameters might be distributed differently in the same layer. However, simply performing incremental training on the sparse network will only update the value of nonzero parameters but keeps the sparse structure fixed.
\end{itemize}

Empirically, we find that naively applying incremental training on the sparse models is able to adapt to the data distribution shift \emph{only in a short period} as the data distribution only shifts a little bit in such a short period. We refer readers to Section~\ref{sec: s2s hard} for detailed empirical results.

\subsection{Capture Data Distribution Shift with Full Model and then Prune} \label{sec: incremental_data_shift}

To address the difficulties of adapting the sparse model to capture the data distribution shift, we propose the Dense-to-Sparse (D2S) paradigm. It maintains a full model and applies incremental training on it to adapt to the data distribution shift. The sparse model is then produced periodically from the maintained full model using a customized pruning algorithm.

\begin{figure*}[ht]
    \centering
    \includegraphics[width=0.8\textwidth]{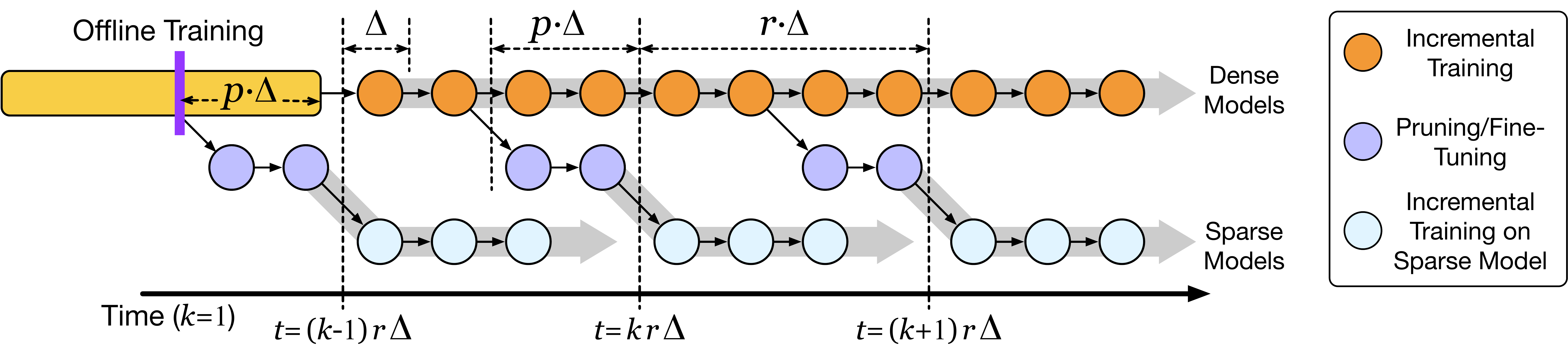}
    \vspace{-5mm}
    \caption{Overview of the Dense-to-Sparse paradigm where the dense model is periodically pruned to produce a sparse model.}\label{fig:fast_branch_aux}
    \vspace{-4mm}
\end{figure*}

Figure~\ref{fig:fast_branch_aux} shows an overview of the D2S paradigm. For every period of $r\Delta$, we will generate a sparse model with updated sparsity structures, learned from the recent full model with the latest data. As an example, at time $t=kr\Delta$, we will generate a new sparse model $f^{sparse}_{\th_{kr\Delta}}$ by pruning the full model $f_{\th_{(kr-p)\Delta}}$ at time $t=(kr-p)\Delta$. The incremental training data in the time window $[(kr-p)\Delta, kr\Delta)$ is used for pruning and necessary fine-tuning.

The learned sparse model will be deployed for serving during the time window $[kr\Delta, (k+1)r\Delta)$. Notice that in this period, the incremental training will also be applied to the sparse model. It means, at $t \in \mathcal{T}\cap[kr\Delta, (k+1)r\Delta) = \left\{(kr+1)\Delta, (kr+2)\Delta, ..., ((k+1)r-1)\Delta\right\}$, the sparse model is incrementally trained using the training data between $(t-\Delta)$ and $t$.

The hyperparameter $r$ needs to be carefully chosen for the D2S paradigm. In an ideal case, we need to choose the $r$ value to satisfy two requirements:
\begin{itemize}
\item For the time $t\in ((k-1)r\Delta, kr\Delta]$, the loss of the sparse model $f^{sparse}_{\th_{(k-1)r\Delta}}$ needs to be lower than or similar to a new sparse model generted by pruning the full model $f_{\th_{t-p\Delta}}$.
\item For the time $t\in \big[kr\Delta, T]$, the loss of the sparse model $f^{sparse}_{\th_{(k-1)r\Delta}}$ will be higher than a new sparse model generated by pruning the full model $f_{\th_{t-p\Delta}}$.
\end{itemize}
The $r$ value cannot be larger because the data distribution shift will hurt the accuracy of the stale sparse models. On the other hand, considering the noise and variation in the incremental training, we can conservatively choose a slightly lower $r$ value. However, $r$ cannot be too small. As shown in Figure~\ref{fig:fast_branch_aux}, for the time window $[(kr-p)\Delta, kr\Delta)$, we need to launch three jobs for the incremental training (dense and sparse models) and pruning/fine-tuning. With a smaller $r$ value, more training resources will be spent on pruning/fine-tuning. In the worst case, if $r=p$, we need to maintain $(p+1)$ jobs in parallel, for which the requirement of training resources is unacceptable.

\subsection{Requirements for Pruning Algorithm}
The special design of D2S raises several requirements for the pruning algorithm. First, D2S requires a pruning algorithm that produces a sparse model given a well-trained full model, and thus, only the pruning after training algorithm is desirable in our system.

Second, following the discussion in Section~\ref{sec: incremental_data_shift} of the choice of $r$, if the sparse model obtained by the pruning algorithm is able to recover its original performance with only a small number of data, then we are able to choose a smaller $r$, which makes the pruned model suffer less from the data distribution shift issue.

Third, due to the heterogeneous architecture in the recommendation model, it is expected that the optimal sparsity for different layers varies. However, hand-crafted tuning of layer-wise sparsity is untenable due to the high cost. Moreover the layer-wise sparsity can be different at different times. It is desirable to let the algorithm learns the sparsity of each layer.


In summary, a desirable pruning algorithm for online recommendation systems should be a pruning after training algorithm that produces a sparse network which only needs a few data for fine-tuning and learns the sparsity for each layer automatically.

\section{Auxiliary Mask Pruning Algorithm}
Classic pruning after training algorithms use heuristic based criteria to measure the importance of weights and prune the less important weights. Some popular heuristics are magnitude based criteria~\citep{li2016pruning, li2018optimization, park2020lookahead} and Taylor approximation based criteria~\citep{molchanov2016pruning, molchanov2019importance}.

Despite impressive empirical achievement of these heuristic based methods, they require significant amount of fine-tuning in terms of more data or more number of epochs. However, in the presence of data distribution shift, these methods are unable to learn a good sparse network, especially when the amount of data available for fine-tuning is limited. We argue that there are three aspects of these algorithms that can be improved:

(1) Using a heuristic criterion for pruning requires that we set the desired sparsity for each layer. However, instead of setting the sparsity of each layer, which tends to lead to a sub-optimal choice of the sparsity of each layer, it is better to learn the sparsity for each layer.

(2) Those heuristic methods usually require gradual pruning schedules to achieve the best accuracy, which is usually sensitive to hyper-parameters. Instead, there needs to be a principled method that can produce sparse networks with minimal tuning.


(3) Previous work only considers using a single heuristic to measure weight importance. However, we find that combining different heuristics can significantly improve pruning quality. 

Based on these motivations, we propose an auxiliary mask based pruning method. To improve (1) and (2), our algorithm borrows the pruning dynamics of sparse constraint based pruning methods (i.e., \citet{he2017channel, liu2017learning}), and is able to automatically decide the sparsity of each layer and percentage of weight to prune at each iteration. To improve (3), we draw an equivalence between the sparse constraint based method and Taylor approximation based heuristics. We find that our dynamics can be viewed as a fine-grain version of Taylor approximation pruning. Based on this insight, we introduce the idea of using multiple heuristics for pruning. 

Recall that given a well trained dense network $f_{\th}(\x)$ with parameter $\th$, our goal is to reduce the size of the network via constraining $\left\Vert \th\right\Vert _{0}\le c$
given some constraint $c$. We consider adding a binary mask with latent continuous parameters into the network and reduce the size of parameters by obtaining sparse masks. That is, for parameter $\theta_{i}$, we equip it with a binary mask with a latent continuous parameter $a_{i}$ and produced a masked parameter $\mathbb{I}\{a_{i}>0\}\theta_{i}$. We are able to prune $\theta_{i}$ out by letting $a_{i}\le0$ and vice versa. In this way, by learning sparse masks (allowing most $a_{i}$ parameters to be non-positive), we are able to zero out a large portion of the parameters. An advantage of using binary mask is that that the algorithm will not over-shrink the unpruned weights.

We consider the following problem
\[
\min_{\a}\mathcal{L}\left[f_{\mathbb{I}(\a>0)\circ\th}\right]\ \ \text{s.t.}\ \ \left\Vert \mathbb{I}(\a>0)\right\Vert _{0}=\left\Vert \mathbb{I}(\a>0)\right\Vert _{1}\le c.
\]
Here $\circ$ denotes the element-wise production. Notice that since we consider binary mask, $\left\Vert \mathbb{I}(\a>0)\right\Vert _{0}=\left\Vert \mathbb{I}(\a>0)\right\Vert _{1}$. It is equivalent to consider training $\a$ by minimizing the following penalized loss 
\[
\min_{\a}\mathcal{L}\left[f_{\mathbb{I}(\a>0)\circ\th}\right]+\lambda\left\Vert \mathbb{I}(\a>0)\right\Vert _{1}.
\]
Here $\lambda$ is the penalty term that enforces the sparsity, and higher $\lambda$ gives higher sparsity. Notice that as the indicator function $\mathbb{I}(\a)$ is not differentiable and to overcome, we use the straight-through estimator~\citep{bengio2013estimating,hinton2012neural}, which replaces the ill-defined gradient of a non-differentiable function in the chain rule by a fake gradient. Specifically, we consider the following update rule at iteration $k$ with learning rate $\epsilon_k$
\begin{align}\label{equ: update original}
\nonumber
\a_{k+1} & =\a_{k}-\epsilon_{k}\nabla_{\left(\mathbb{I}(\a>0)\right)}\mathcal{L}_{\D}\left[f_{\mathbb{I}(\a>0)\circ\th}\right]q(\a)
\\
& -\epsilon_{k}\lambda\nabla_{\left(\mathbb{I}(\a>0)\right)}\left\Vert \mathbb{I}(\a>0)\right\Vert _{0}q(\a)
\\
\nonumber
 & =\a_{k}-\epsilon_{k}\nabla_{\left(\mathbb{I}(\a>0)\right)}\mathcal{L}_{\D}\left[f_{\mathbb{I}(\a>0)\circ\th}\right]q(\a)-\epsilon_{k}\lambda q(\a).
\end{align}
Here $q(\a)$ is the fake gradient used to replace the ill-defined $\nabla_{\a}\mathbb{I}(\a>0)$. Some common choice are $q(\a)=\nabla_{\a}\text{id}(\a)=1$ (Linear STE) or $q(\a)=\nabla_{\a}\text{ReLU}(\a)$ (ReLU STE). Notice that for ReLU STE, the gradients of $\a$ with negative values will always equal to zero. It means once a weight is being pruned, there is no chance for it to be alive again, which makes it impossible for the algorithm to correct the mistake if it mistakenly prunes out some important weights. Based on this intuition, we choose to use Linear STE in this paper.

\paragraph{The Connection with Taylor Approximation Based Pruning}

Choosing $q(\a)=1$ and using the first-order Taylor approximation, we have 
\begin{align*}
\mathcal{L}_{\D}\left[f_{\th}\right]-\mathcal{L}_{\D}\left[f_{\th_{-i}}\right] & \approx\nabla_{\left(\mathbb{I}(a_{i}>0)\right)}\mathcal{L}_{\D}\left[f_{\mathbb{I}(\a>0)\circ\th}\right]\\
 & =\nabla_{\left(\mathbb{I}(a_{i}>0)\circ\theta_{i}\right)}\mathcal{L}_{\D}\left[f_{\mathbb{I}(\a>0)\circ\th}\right]\theta_{i}\;\;.
\end{align*}
Here $\th_{-i}$ denotes the parameter with $\theta_i$ set to 0, and all other elements remain the same as $\th$.
From this perspective, our algorithm calculates the weight importance (with batches of data) using the first-order Taylor approximation. Once there is enough evidence showing that a certain weight is not important by finding that its corresponding mask becomes zero, this weight is pruned.

\paragraph{Combining Multiple Criteria: the Generalized Pruning Dynamics}

Notice that our algorithm is able to: 1) learn the sparsity of each layer; 2) do iterative pruning with pruning percentage automatically decided. To further improve the result, we find that combining multiple criteria is better than simply using one criterion to measure weight importance. We make the following modification over the vanilla version.

Firstly, we find that it is more useful to measure the importance of weight by how much the loss changes (i.e., $|\mathcal{L}_{\D}\left[f_{\th}\right]-\mathcal{L}_{\D}\left[f_{\th_{-i}}\right]|$) if pruned rather than how much the loss decreases (i.e., $\mathcal{L}_{\D}\left[f_{\th}\right]-\mathcal{L}_{\D}\left[f_{\th_{-i}}\right]$). This is consistent with the findings in computer vision \citep{molchanov2016pruning, molchanov2019importance} and can be achieved by using the absolute value of Taylor approximation.

Besides, we find it useful to also include weight magnitude information to measure weight importance. Incorporating all the motivations, we consider the following update rule 
\begin{eqnarray} \label{equ: update no rescale}
\a_{k+1}=\a_{k}-\epsilon_{k}w_1g_{1}(\a)-\epsilon_{k}w_2g_{2}(\a)-\epsilon_{k}\lambda,
\end{eqnarray}
where we have
\begin{align*}
g_{1}(a_{i}) =-\left|\nabla_{\left(\mathbb{I}(a_{i}>0)\right)}\mathcal{L}_{\D}\left[f_{\mathbb{I}(\a>0)\circ\th}\right]\theta_i\right|,
\ \ 
g_{2}(a_{i}) =-|\theta_{i}|.
\end{align*}
Here $w_1, w_2$ control the ratio of the information coming from Taylor approximation and weight magnitude. Notice that here we are no more solving an optimization problem but doing a fine-grain version of weight importance measurement.

\vspace{-0.2cm}
\paragraph{The Gradient Rescaling Trick}
Directly applying the update rule in Equation~\ref{equ: update no rescale} is problematic. The reason is that the Taylor Approximation term and weight magnitude term may have a very different scale for each layer and across different layers. To resolve this issue, we proposed the gradient rescaling trick, in which we normalize each component of the gradient to ensure they have the same scale. The final update rule is as follows:
\begin{eqnarray} \label{equ: final update}
\a_{k+1}=\a_{k}-\epsilon_{k}w_1\bar{g}_{1}(\a)-\epsilon_{k}w_2\bar{g}_{2}(\a)-\epsilon_{k}\lambda,
\end{eqnarray}
where we have
\begin{align*}
\bar{g}_{1}(a_{i}) & =-\frac{\left|\nabla_{\left(\mathbb{I}(a_{i}>0)\right)}\mathcal{L}_{\D}\left[f_{\mathbb{I}(\a>0)\circ\th}\right]\theta_{i}\right|}{\left\Vert \nabla_{\left(\mathbb{I}(\a>0)\right)}\mathcal{L}_{\D}\left[f_{\mathbb{I}(\a>0)\circ\th}\right]\circ\th\right\Vert _{1}}\\
\bar{g}_{2}(a_{i}) & =-\frac{\left|\theta_{i}\right|}{\left\Vert \th\right\Vert _{1}}.
\end{align*}
Here $||\cdot||$ denotes the vector $\ell_1$ norm. We find that without using the weight magnitude information, the error increases about 120\% and if we directly use the vanilla updating rule in Equation~\ref{equ: update original}, the error increases about 2400\%.

\begin{figure*}[t] 
    \centering
    \includegraphics[width=0.67\textwidth]{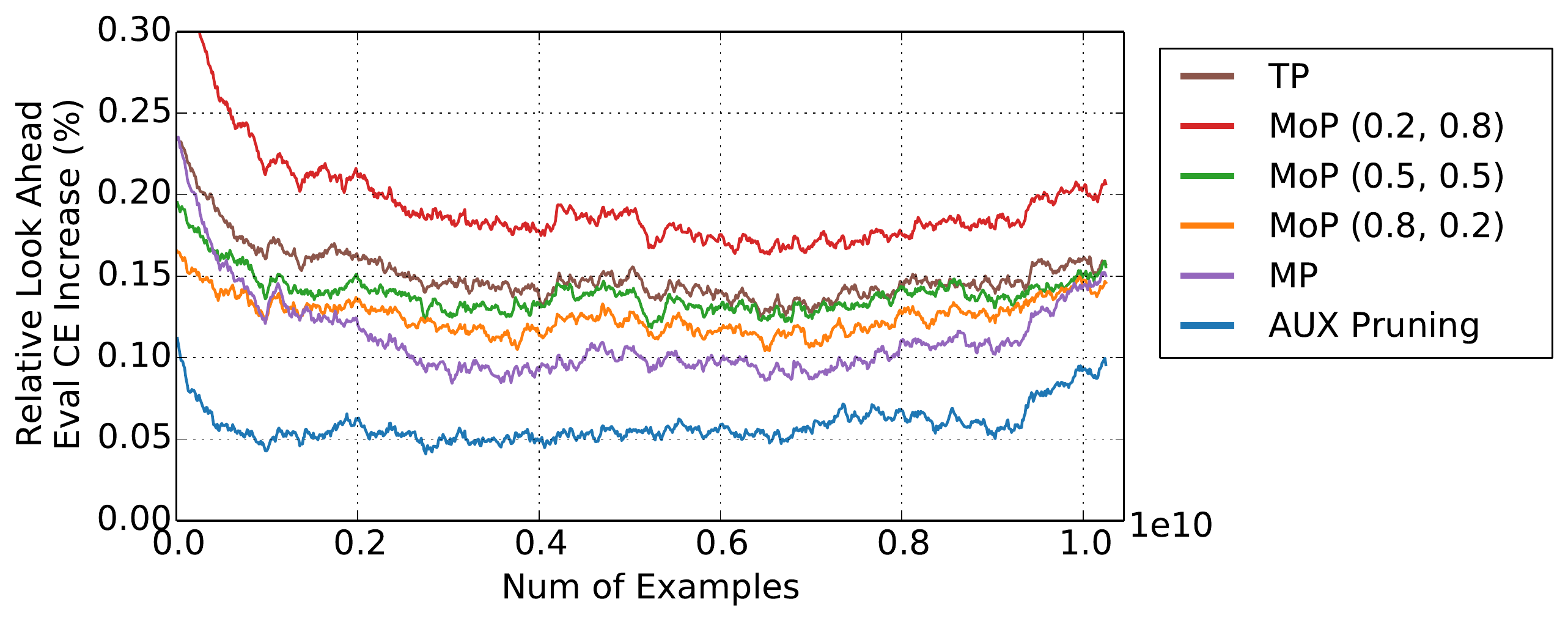}
    \vspace{-6mm}
    \caption{Pruning algorithms comparison based on the look ahead window CE relative to the full dense model. \textbf{MP}: Magnitude Pruning; \textbf{TP}: Taylor Approximation Pruning; \textbf{MoP $(w_1,w_2)$}: Pruning based on different weighted combinations of Magnitude and Momentum.}
    \label{fig:techniques}\label{fig: all}
    \vspace{-5mm}
\end{figure*}

\section{Experiments}
\paragraph{Training Setup}
We conduct our numerical experiments on training recommendation models for click-through rate prediction tasks on large scale systems. We use the DLRM model architecture \citep{naumov2019deep} to design the networks with both dense and categorical features and use the dot-product interaction method. Please note that the proposed methods are agnostic to the model architecture, and are applicable to most recommendation models.

The models are trained using an asynchronous data-parallel distributed setup, where we have multiple trainer nodes working on a chunk of the data, and a centralized server that synchronizes the weights across trainers. Similar to \citet{zheng2020shadowsync}, we perform a single pass on the training data. We use the Adagrad \citep{adagrad} optimizer for training the network parameters, and all hyper-parameters are selected using cross-validation. We use pre-trained models trained with $50B$ data samples, and then train them incrementally for $1B$ data samples for all experiments. 

We use the stochastic gradient descent optimizer for learning the latent auxiliary parameter $\boldsymbol{a}$ of the mask. During the learning of the sparse mask, we also allow the (unpruned) weights to be updated. Unpruned weights are the ones for whom the auxiliary parameter $\boldsymbol{a}$ is greater than zero. For the proposed algorithm, we achieve the targeted sparsity by tuning the penalty $\lambda$, which penalizes $\boldsymbol{a}$.
For example, $\lambda = 1.75$ and $\lambda = 1.875$ gives around 0.75 (75\% of weights are zeroed out) and 0.80 overall sparsity, respectively. We also note that the prune penalty is stable and it produces the same desired sparsity for different $k$ values (length of the pruning phase). This is very important to avoid re-tuning $\lambda$ for different data segments.
We set $w_1=w_2=0.5$, giving equal weights to the information coming from Taylor approximation and weight magnitude terms for AUX. We aim to achieve over-all target sparsity ratios of 0.8, which is expected to give up to 2x speedup on large scale models.

\vspace{-0.2cm}
\paragraph{Evaluation}
For model accuracy, we report the cross-entropy loss (CE) from the classification task. More specifically, recommendation systems often use the Normalized Cross Entropy~\citep{predicting-ne} metric which measures the average log loss divided by what the average log loss would be if a model predicted the background click-through rate.

Given that we care more about the performance of incoming new data rather than historical performance over the whole dataset, we also consider the look ahead window evaluation (eval) CE \citet{cervantes2018evaluating}, in which the CE is calculated using the data (before passing it to the optimizer) within the moving time window (e.g., $10M$ samples for this paper). Since we only pass each data into the optimizer once and the data used for calculating the look ahead CE has not been passed to the optimizer at the time of evaluation, the look ahead eval CE is an improved version of the  windowed evaluation loss. We report the relative CE of the compared methods. The relative CE is defined by
\vspace{-0.2cm}
\[
\text{relative\ CE} = \frac{\text{CE\ of\ the\ pruned\ model}}{\text{CE\ of\ the\ full\ model}}-1.
\]
\vspace{-0.6cm}
\paragraph{Compared Methods}
We compare the baselines below with the proposed Auxiliary Mask Pruning method (AUX). 

(1) Magnitude based pruning (MP)~\citep{li2016pruning}, which uses weight magnitude to rank the weight importance and prunes the least important weight. We consider an iterative version of this magnitude based pruning, in which we gradually and linearly increase the pruning ratio (i.e., the percentage of pruned weight) from zero to targeted sparsity over a period of time (pruning phase).

(2) Taylor Approximation based Pruning (TP)~\citep{molchanov2016pruning} is similar to MP but measure the weight importance via (first-order) Taylor approximation.

(3) Momentum Based Pruning (MoP)~\citep{ding2019global, dettmers2019sparse, evci2020rigging}: Notice that as we have non-stationary data and thus it is also reasonable to measure the weight importance by its momentum (calculated with the exponential moving average of gradients). Larger momentum means the weight is more important for recent data, as its magnitude of updates is larger. However, empirically, we find that naively applying those methods gives very poor results. We believe this is because some of the very high magnitude weights (which may have low momentum) affect the network a lot when pruned. We enhance momentum techniques via measuring the importance of weight with both (normalized) weight magnitude and (normalized) momentum magnitude. That is, the weight importance $\theta_\text{imp}$ is calculated by
\vspace{-0.1cm}
\[
\theta_\text{imp} = w_1|\theta|/||\th||_1 + w_2|s|/||\boldsymbol{S}||_1,
\]
where $s$ is the momentum of the weight and $\boldsymbol{S}$ is the momentum of all the weights in the same layer. Momentum $S$ is calculated using the decay parameter $0.99$ (chosen using cross-validation). This algorithm does not prune weights with large magnitude, but at the same time keeps weights with low magnitude and high momentum. This achieves the same objective as methods like Rigged Lottery \citep{evci2020rigging}, Sparse Networks From Scratch \citep{dettmers2019sparse}, which periodically prune and grow weights based on gradient momentum.

\begin{figure*}[t]
\center
\includegraphics[width=0.75\textwidth]{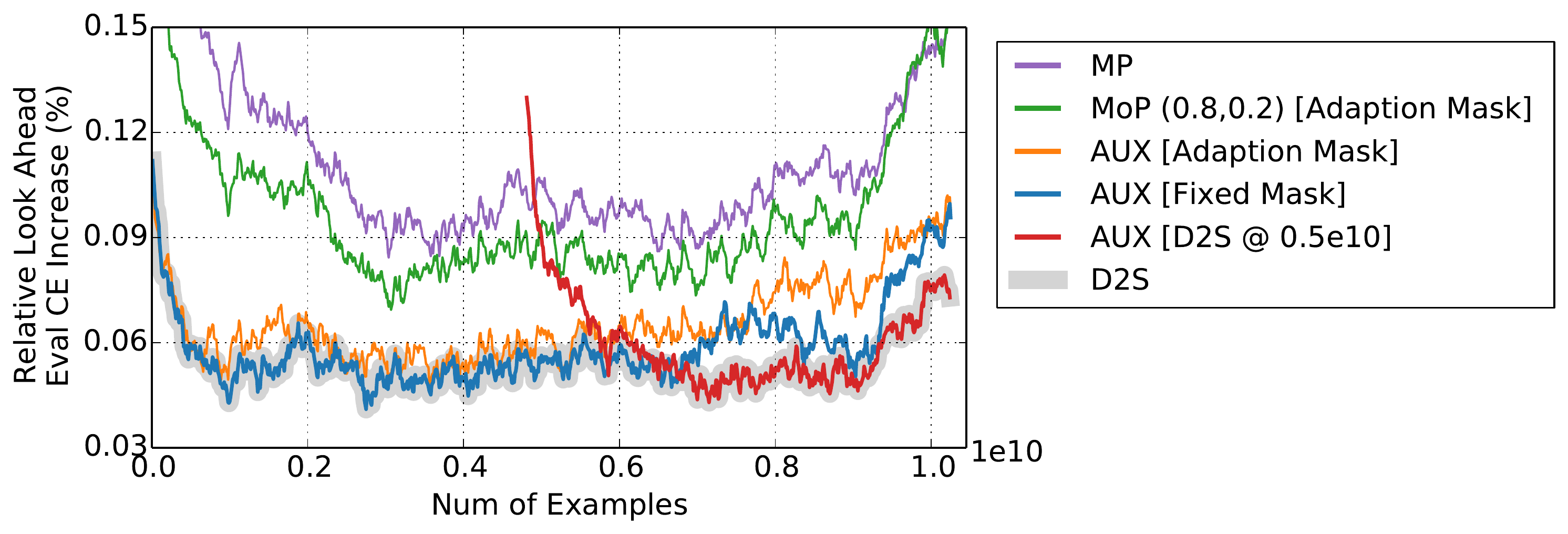}
\vspace{-6mm}
\caption{Mask Adaptation to capture data non-stationarity. \textbf{MP}: Magnitude Pruning with fixed mask; \textbf{MoP (0.8, 0.2)}: Momentum based pruning with $w_1=0.8$ and $w_2=0.2$; \textbf{AUX [Adaptation Mask]}: AUX pruning with no fine-tuning phase; \textbf{AUX [Fixed Mask]}: AUX pruning with pruning and then fine-tuning with fixed mask; \textbf{AUX [D2S @ 0.5e10]}: AUX pruning that is applied on the dense model at 0.5e10 samples; \textbf{D2S}: The best of AUX pruning based sparse networks from AUX [Fixed Mask] \& AUX [D2S @ 0.5e10].}\label{fig:s2s}
\vspace{-3mm}
\end{figure*}

\subsection{Pruning with Fixed Mask}
We first evaluate the performance of the pruning algorithm itself independent of the impact of data non-stationarity. We start with a pre-trained model using up to $50B$ samples, apply the pruning phase on $100M$ samples, and then fix the sparsity structure during subsequent incremental training phases (only allowing the unpruned weights to be updated). Empirically we found that $100M$ samples are sufficient to learn the mask for all the methods, and increasing this does not improve the accuracy.  Figure~\ref{fig: all} shows the results based on the relative window look ahead eval CE. AUX outperforms all the baselines with a large margin. Interestingly, all momentum based techniques are worse than pure magnitude based pruning (MP). This shows that MP is a very strong baseline to begin with, which has been shown previously as well~\citep{kusupati2020soft}. Moreover it is quite challenging to augment the magnitude information with momentum information. 

Figure~\ref{fig: all} also shows that the AUX method reaches steady state loss much before other methods. For instance, the AUX method reaches steady state after $0.1e10$ samples whereas MP requires at least $0.3e10$ samples. This also shows that AUX pruning does a much better job of removing unnecessary weights and causes much less disruption to the learning process. Table~\ref{table: all} also gives the relative window look ahead eval CE of the latest $10M$ samples and eval CE calculated using the $1B$ samples right after training ends. 

\begin{table}[h]
\small
\begin{centering}
\begin{tabular}{c|cc}
\toprule[0.75pt]
Method & Look Ahead Eval CE  \% & Eval CE \% \tabularnewline
\hline 
MP & 0.149 & 0.191\tabularnewline
TP & 0.157 & 0.184\tabularnewline
MoP (0.5, 0.5) & 0.156 & 0.211\tabularnewline
MoP (0.2, 0.8) & 0.206 & 0.246\tabularnewline
MoP (0.8, 0.2) & 0.147 & 0.254\tabularnewline
AUX & \textbf{0.097} & \textbf{0.144}\tabularnewline
\bottomrule[0.75pt]
\end{tabular}
\vspace{-3mm}
\caption{Increase in the last-window look ahead eval CE and the eval CE of the compared methods. We try different weight magnitude and momentum combinations denoted as MoP $(w_1,w_2)$.}
\label{table: all}
\par\end{centering} 
\vspace{-5mm}
\end{table}

\subsection{Difficulty of Directly Adjusting Sparse Network for Non-stationary Data} \label{sec: s2s hard}

We empirically show that adapting the mask to changing distribution is a non-trivial problem and all pruning methods suffer from this. We give a more extensive study on several potential mask adaption methods for adjusting the sparse model to adapt to the data non-stationarity. 
\vspace{-4mm}
\paragraph{Mask Adaption Methods}
(1) AUX (Fixed Mask): This is the baseline scheme where we apply AUX pruning for $100M$ samples and then continue to fine-tune the fixed mask for the rest of the training. This is exactly the same way we propose to fine-tune the sparse model during its serving period (i.e., $\left[kr\Delta,(k+1)r\Delta\right]$).

(2) AUX (Adaptation Mask): This is a straightforward way to extend AUX pruning for mask adaptation. We continue to update the mask and the weights together continuously throughout incremental training. Due to the fact that we use Linear STE as the fake gradient for the non-differentiable indicator function of the auxiliary mask $a$, weights can continuously be pruned and unpruned leading to a natural adaptation of the mask. There is no explicit fine-tuning phase here, where the auxiliary mask is fixed. It should be noted that the final achieved sparsity depends solely on the regularization strength $\lambda$ and not on the length of the pruning phase. Hence, as long as $\lambda$ is kept constant, the overall achieved sparsity remains stable.

(3) MoP (Adaptation Mask): Similar to AUX (Adaption Mask), we consider continuously updating the mask for MoP using the latest momentum and weight magnitude values. We update the mask using this criterion for every $n$ samples. Using cross-validation we found that optimal value for $n$ is $10M$.  This is similar to many recent pruning during training methods that employ momentum as the importance metrics to rank weights \citep{ding2019global, dettmers2019sparse, evci2020rigging}. This technique takes advantage of the fact that even though the magnitude of pruned weights is not updated, their momentum can still increase, which may allow them to be selected in the next selection cycle.

(4) AUX (Dense-to-Sparse): This method combines AUX with the Dense-to-Sparse paradigm. Periodically the dense model is used to produce a fresh sparse model. We use AUX (D2S @x) to indicate that the second sparse network is generated after x examples. We consider $x=0.5e10$ as the offset at which the second sparse network is instantiated by pruning the full dense network.

Here we only consider modifying the MoP, as MP and TP prune weights in an unrecoverable way and thus cannot be easily modified to allow the sparsity structure to change. For instance, in MP, the pruned weights are never updated, and hence they do not have a chance to be selected again.
\vspace{-4mm}
\paragraph{Mask Adaptation Results} The results are summarized in Figure \ref{fig:s2s}. It can be shown that all the methods that directly update the sparse model are not able to make the sparse network adapt to the new data as the window CE loss start to increase after $0.6e10$ samples. AUX (Adaptation Mask) is also unable to adapt to the distribution shift, which illustrates our claim that under-parameterized networks struggle to adapt irrespective of the pruning algorithm.

In comparison, the proposed AUX (D2S @ 0.5e10) gives much lower CE loss because it depends on the dense network to adapt to the data distribution shift. D2S paradigm is powerful because we can produce sparse networks by pruning the dense networks at any chosen time. For instance, the shaded algorithm D2S illustrates how the two sparse networks generated at $0$ samples and $0.5e10$ samples can be combined to give the best overall accuracy. How often we can produce a sparse network is dependent on how fast can a pruned network converges to steady state. This is where the AUX pruning algorithm is ideal because it converges very fast after pruning is applied.


One difficulty considering the data distribution shift is that it is very difficult to predict when the model accuracy will start to suffer from the stale mask. In Figure~\ref{fig: all}, we can see that the data distribution shift starts to hurt the model accuracy after $0.6e10$ samples. This threshold can vary for different architectures, different data segments, and different target sparsity ratios. A critical advantage of D2S is that we always keep a dense model for comparison. In this case, we can monitor the accuracy difference between the dense and sparse model to determine the $r$ value and dynamically produce sparse models when the accuracy starts to diverge.



\begin{figure}[t]
\center
\includegraphics[width=0.45\textwidth]{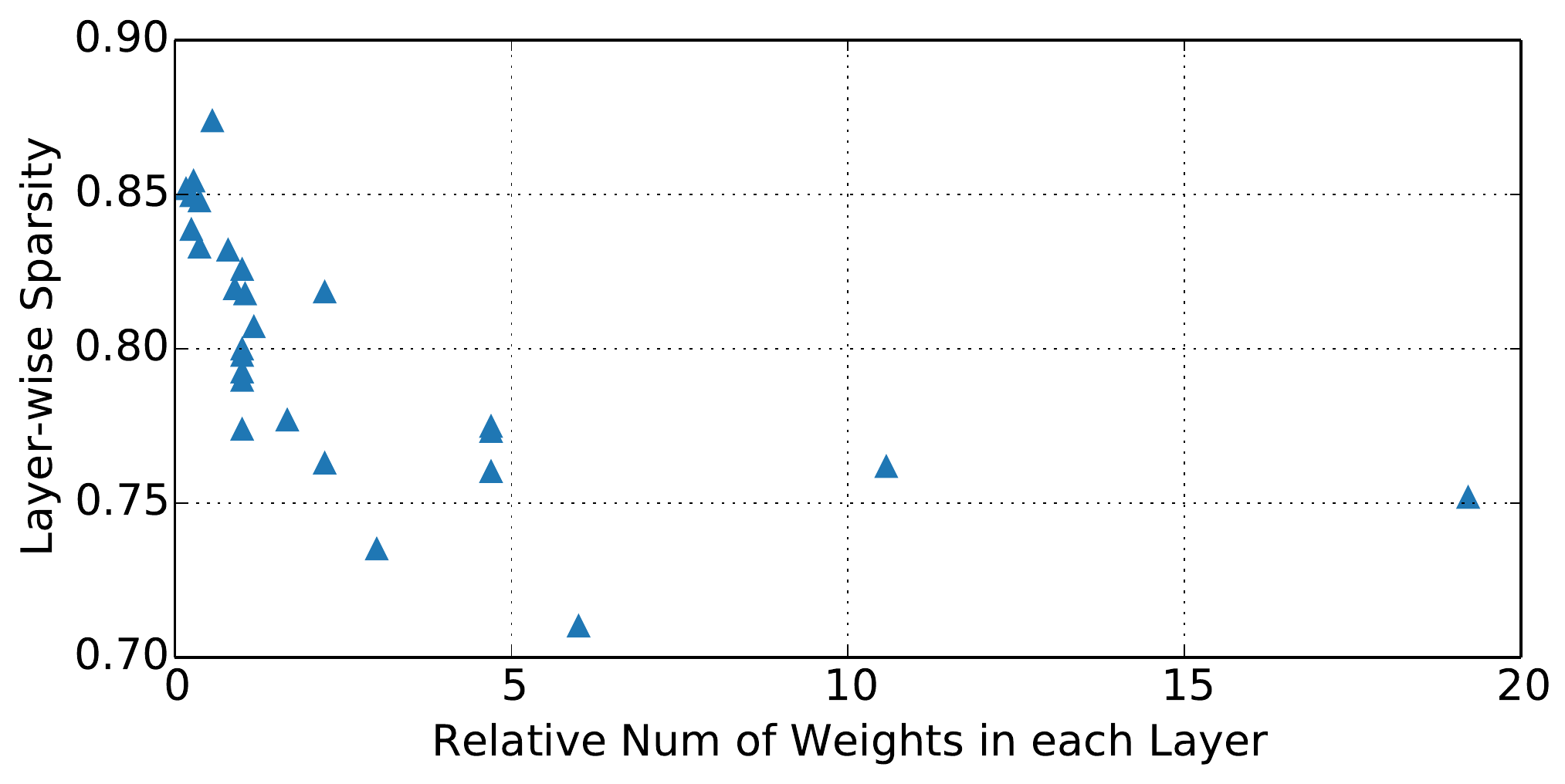}
\vspace{-5mm}
\caption{Layer-wise sparsity vs. the relative number of weights in each layer with respect to the smallest layer.} \label{fig:sparse_vs_size}
\vspace{-5mm}
\end{figure}

\subsection{Learned sparsity for each layer}
AUX pruning automatically learns the sparsity of each layer, which is critical in reducing the overall accuracy loss. Figure \ref{fig:sparse_vs_size} plots the final achieved sparsity across layers plotted against the relative size of the layer. It suggests that the algorithm tends to give larger sparsity to those layers with smaller size. This is due to the heterogeneity in the architecture where the size of the layer does not necessarily dictate the redundancy of the parameters, contrary to popular belief that larger layers are more redundant. Hence applying the same sparsity to all layers or only pruning the largest layers are sub-optimal choices.

\begin{figure}[t]
\center
\includegraphics[width=0.45\textwidth]{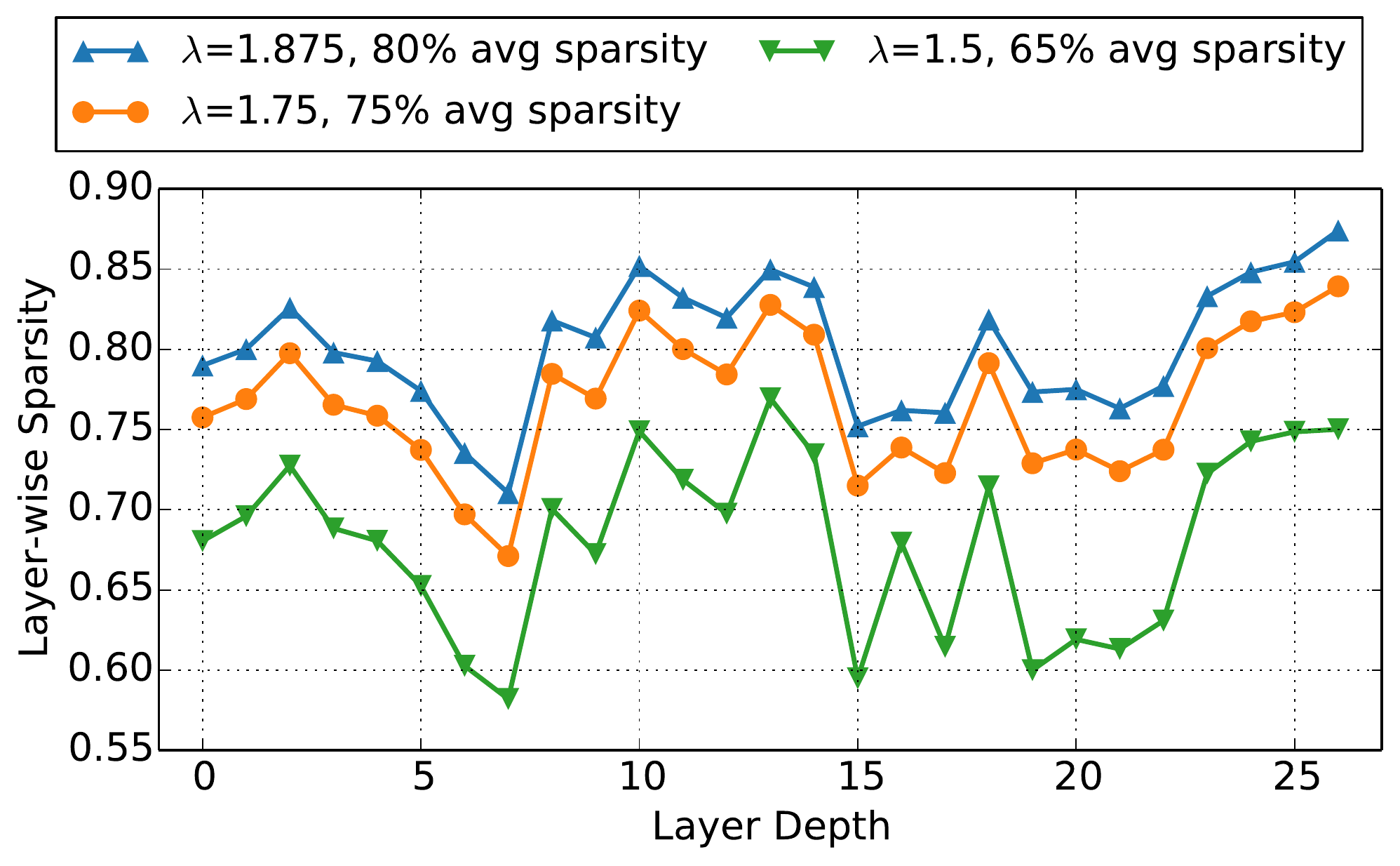}
\vspace{-5mm}
\caption{Layer-wise sparsity vs. layer depth for different regularization strengths ($\lambda$).} \label{fig:sparse_vs_depth}
\vspace{-4mm}
\end{figure}

Figure \ref{fig:sparse_vs_depth} plots the layer-wise sparsity vs. the depth of the layer for three different regularization strengths ($\lambda$), which produce different overall target sparsity ratios. This shows that in contrast to domains like Computer Vision, there is no obvious correlation between sparsity and depth.
This makes it more desirable for the algorithm to learn the sparsity instead of using handcrafted rules. We can also see how the $\lambda$ values can be tuned to modulate the overall sparsity of the network. We also find that a given $\lambda$ value produces consistent sparsity ratios across different date ranges and model sizes, which reduces the overhead of tuning $\lambda$.



\subsection{Significance of Combining Multiple Criteria}
\begin{figure}
\center
\includegraphics[width=0.44\textwidth]{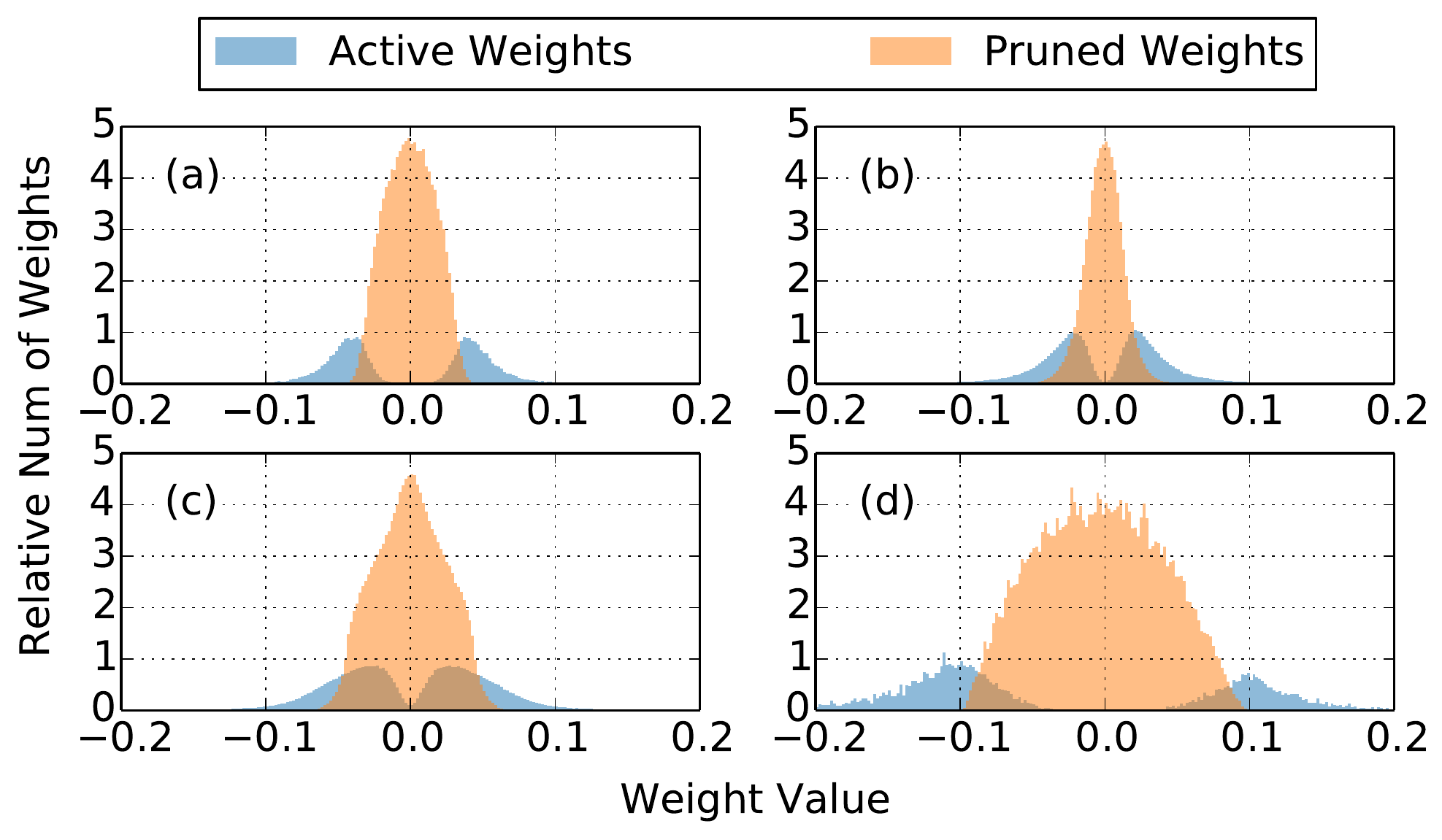}
\vspace{-6mm}
\caption{Histogram of the pruned and active weights for 4 different FC layers.} \label{fig: dist of weight}
\vspace{-5mm}
\end{figure}

Results in Figure \ref{fig: all} suggest that except for the proposed AUX algorithm, MP seems to give the best performance. It is thus of interest to understand whether adding the Taylor approximation term significantly changes the decision on what kind of weights are pruned compared with MP. Figure \ref{fig: dist of weight} plots the histogram of the pruned weights and the unpruned weights in four different FC layers. It shows that the algorithm prunes weight with very small magnitude and keeps weights with very large magnitude. However, for a large portion of weights with `moderate' magnitudes, the algorithm carefully decides their importance by checking the information of the Taylor approximation term, which causes the huge overlap of weight histogram of the two groups. This shows that gradient information is very important in deciding the weight importance for recommendation models. However, simple momentum based ranking heuristics are unable to harness that information. Hence a formal algorithm like AUX pruning outperforms momentum based algorithms.

\section{Conclusion \& Future Work}
We have proposed a novel pruning algorithm coupled with a novel Dense-to-Sparse (D2S) paradigm that is designed for application of pruning to large scale recommendation systems. We have discussed the implications of this algorithm on system design and shown the efficacy of the algorithms on large scale recommendation systems.
D2S is effective in lowering the accuracy loss but is inefficient during training. We need to train a dense model at all times, and thus the number of model replicas is increased during training. In the future, it will become important to improve mask adaptation techniques to reduce the training overhead.

\nocite{langley00}

\bibliography{main}
\bibliographystyle{mlsys2020}



\end{document}